\documentclass[letterpaper]{article} 
\usepackage{aaai2026}  
\usepackage{times}  
\usepackage{helvet}  
\usepackage{courier}  
\usepackage[hyphens]{url}  
\usepackage{graphicx} 
\usepackage{natbib}  
\usepackage{caption} 
\usepackage{algorithm}
\usepackage{algorithmic}

\usepackage{newfloat}
\usepackage{listings}
\DeclareCaptionStyle{ruled}{labelfont=normalfont,labelsep=colon,strut=off} 
\floatstyle{ruled}
\newfloat{listing}{tb}{lst}{}
\floatname{listing}{Listing}
\usepackage{booktabs}
\usepackage{amsmath}
\usepackage{amssymb}
\usepackage{colortbl}
\usepackage{xcolor}
\usepackage{multirow}
\usepackage{placeins}
\usepackage{makecell}

\newcommand{\method}{SpatialActor}

\title{\method: Exploring Disentangled Spatial Representations \\
for Robust Robotic Manipulation}
\author{
    Hao Shi\textsuperscript{\rm 1}\thanks{Work done during internship at Dexmal.},~
    Bin Xie\textsuperscript{\rm 2},~
    Yingfei Liu\textsuperscript{\rm 2},~
    Yang Yue\textsuperscript{\rm 1},~
    Tiancai Wang\textsuperscript{\rm 2},\\
    Haoqiang Fan\textsuperscript{\rm 2},~
    Xiangyu Zhang\textsuperscript{\rm 3,}\textsuperscript{\rm 4},~
    Gao Huang\textsuperscript{\rm 1}\thanks{Corresponding author: Gao Huang.}
}
\affiliations{
    \textsuperscript{\rm 1}Department of Automation, BNRist, Tsinghua University \\
    \textsuperscript{\rm 2}Dexmal\\
    \textsuperscript{\rm 3}MEGVII Technology\\
    \textsuperscript{\rm 4}StepFun \\
    
    shi-h23@mails.tsinghua.edu.cn, wtc@dexmal.com, gaohuang@tsinghua.edu.cn
}

\usepackage{bibentry}

\begin{document}

\maketitle

\begin{abstract}

Robotic manipulation requires precise spatial understanding to interact with objects in the real world. 
Point-based methods suffer from sparse sampling, leading to the loss of fine-grained semantics. 
Image-based methods typically feed RGB and depth into 2D backbones pre-trained on 3D auxiliary tasks, but their entangled semantics and geometry are sensitive to inherent depth noise in real-world that disrupts semantic understanding. 
Moreover, these methods focus on high-level geometry while overlooking low-level spatial cues essential for precise interaction. 
We propose \textit{\method}, a disentangled framework for robust robotic manipulation that explicitly decouples semantics and geometry. 
The Semantic-guided Geometric Module adaptively fuses two complementary geometry from noisy depth and semantic-guided expert priors. 
Also, a Spatial Transformer leverages low-level spatial cues for accurate 2D-3D mapping and enables interaction among spatial features. 
We evaluate \textit{\method} on multiple simulation and real-world scenarios across  50+ tasks. 
It achieves state-of-the-art performance with 87.4\% on RLBench and improves by 13.9\% to 19.4\% under varying noisy conditions, showing strong robustness. 
Moreover, it significantly enhances few-shot generalization to new tasks and maintains robustness under various spatial perturbations. 
\end{abstract}

\begin{links}
\link{Project Page}{https://shihao1895.github.io/SpatialActor}
\link{Code}{https://github.com/shihao1895/SpatialActor}
\end{links}

\section{Introduction}
\label{sec:intro}

\begin{figure*}[ht]
\hsize=\textwidth
    \centering
    \includegraphics[width=1\linewidth]{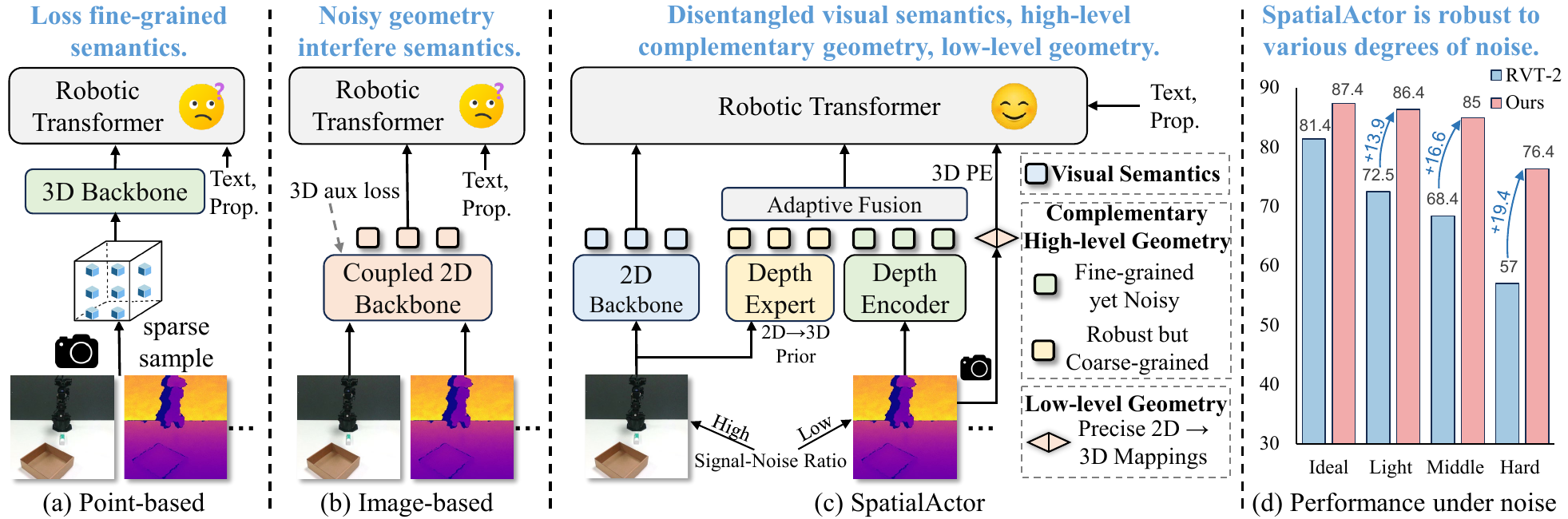}
    \caption{\textbf{Methodology comparisons}. 
    (a) Point-based methods suffer from sparse sampling, leading to the loss of fine-grained semantics. 
    (b) Image-based methods typically entangle semantics and geometry, while inherent depth noise in real-world disrupts semantic understanding. 
    (c) \method~disentangle visual semantics, two complementary high-level geometry from noisy depth and expert priors, low-level spatial cues. 
    (d) Performance under various degrees of noise, showing the robustness. }
    \label{fig:intro}
\end{figure*}

Robotic manipulation enables robots to understand scenes and interact with objects to perform precise physical tasks in the real-world environments. 
Some existing methods~\cite{zeng2021transporter,zhao2023learning,brohan2022rt,kim2024openvla,chi2023diffusion,liu2024rdt,shi2025memoryvla} rely solely on 2D visual inputs to predict end-effector actions in 3D space, however, they often struggle in scenarios requiring spatial reasoning, occlusion handling, geometric shape comprehension, or fine-grained object interactions due to their limited understanding of spatial geometry. 
Given that real-world tasks inherently occur in 3D space, incorporating 3D spatial information is crucial for learning robust and generalizable robotic manipulation policies. 

Recent efforts in robotic manipulation have explored various approaches to exploit spatial information. 
In Fig.~\ref{fig:intro} (a), point cloud-based approaches~\cite{zhang2023universal,chen2023polarnet,ze20243d,james2022coarse} represent 3D geometry explicitly, yet suffer from semantic loss due to sparse sampling and are limited by the high cost of 3D annotations, which constrains pretraining scalability. 
In contrast, Fig.~\ref{fig:intro} (b) illustrates image-based methods~\cite{goyal2023rvt,goyal2024rvt,fang2025sam2act,wang2024vihe} that utilize multi-view RGB-D to jointly model semantics and geometry in a shared feature space. 
These methods exploit structured 2D inputs to obtain dense semantics and benefit from strong 2D pretrained priors, enabling competitive performance. 
However, the entanglement of semantics and geometry makes these methods sensitive to inherent depth noise in the real-world, which degrades semantic and geometric understanding. 
As shown in Fig. \ref{fig:intro}~(d), even minor noise can lead to a significant performance drop of 8.9\% in RVT2~\cite{goyal2024rvt}. 
In reality, depth is often compromised by sensor noise, lighting variations, and surface reflections, which severely limit the practical application of such methods in the real-world. 
Furthermore, the joint modeling primarily retains high-level geometry while neglecting low-level spatial cues that are critical for precise interaction by providing fine-grained 2D-3D correspondences. 

The limitations above call for three critical capabilities in robotic manipulation: 1) fine-grained spatial understanding to enable accurate control; 2) robustness to sensor noise to ensure real-world reliability; and 3) low-level spatial cues to support consistent spatial tokens interaction. 
This raises a fundamental question: How can we construct a robust spatial representation that fulfills these requirements?

To address this, we propose \textit{\method}, a novel framework for robust spatial representation in robotic manipulation. 
Instead of a shared feature space, we decouple semantics and geometry to mitigate cross-modal interference. 
Furthermore, we decompose geometric information into high-level geometric representations and low-level spatial cues. 
To construct a robust high-level geometric representation, we propose a Semantic-guided Geometric Module (SGM). 
Within the SGM, high signal-to-noise semantics from RGB are processed by a large-scale pretrained depth estimation expert~\cite{yang2024depth,yang2025depth} to produce a robust but coarse geometric prior. 
Meanwhile, raw depth inputs retain fine-grained geometric details but are inherently noisy. 
By adaptively integrating these complementary geometric representations through a gating mechanism, the SGM enhances both robustness and spatial precision, effectively addressing the limitations of individual modalities. 
For low-level positional cues, we introduce a Spatial Transformer (SPT) that integrates spatial modeling into the transformer layers. 
By employing spatial position encoding, distinct spatial tokens are endowed with unique spatial indices, facilitating spatial interactions. 
The model performs view-level interaction to refine token relationships within each view, followed by scene-level interaction that unifies cross-modal cues across the scene, yielding features for the action head. 

To comprehensively evaluate our method, \method, we conduct experiments on 50+ robotic manipulation tasks in both simulation and real-world. 
In RLBench (18 tasks with 249 variations), \method~achieves an 87.4\% average success rate, surpassing state-of-the-art methods by approximately 6.0\%, with a notable 53.3\% improvement in high-precision spatial tasks like Insert Peg. 
Our method also shows strong robustness, maintaining higher success rates under noise conditions with improvements of 13.9\%, 16.9\%, and 19.4\% at light, medium, and heavy noise levels, respectively. 
On ColosseumBench, which evaluates 20 tasks under spatial perturbations, \method~consistently outperforms baselines, showcasing superior spatial generalization. 
Additionally, in a few-shot setting, adapting a multi-task pre-trained model to 19 novel tasks with only 10 demonstrations per task, \method~achieves 79.2\% success compared to 46.9\% for RVT-2. 
Real-world experiments further validate these results, as \method~outperforms RVT-2 across 8 tasks and 15 variations, demonstrating its strong robustness and generalization across diverse scenarios. 

\section{Related Works}
\label{sec:related}

\begin{figure*}[h]
    \centering
    \includegraphics[width=1\linewidth]{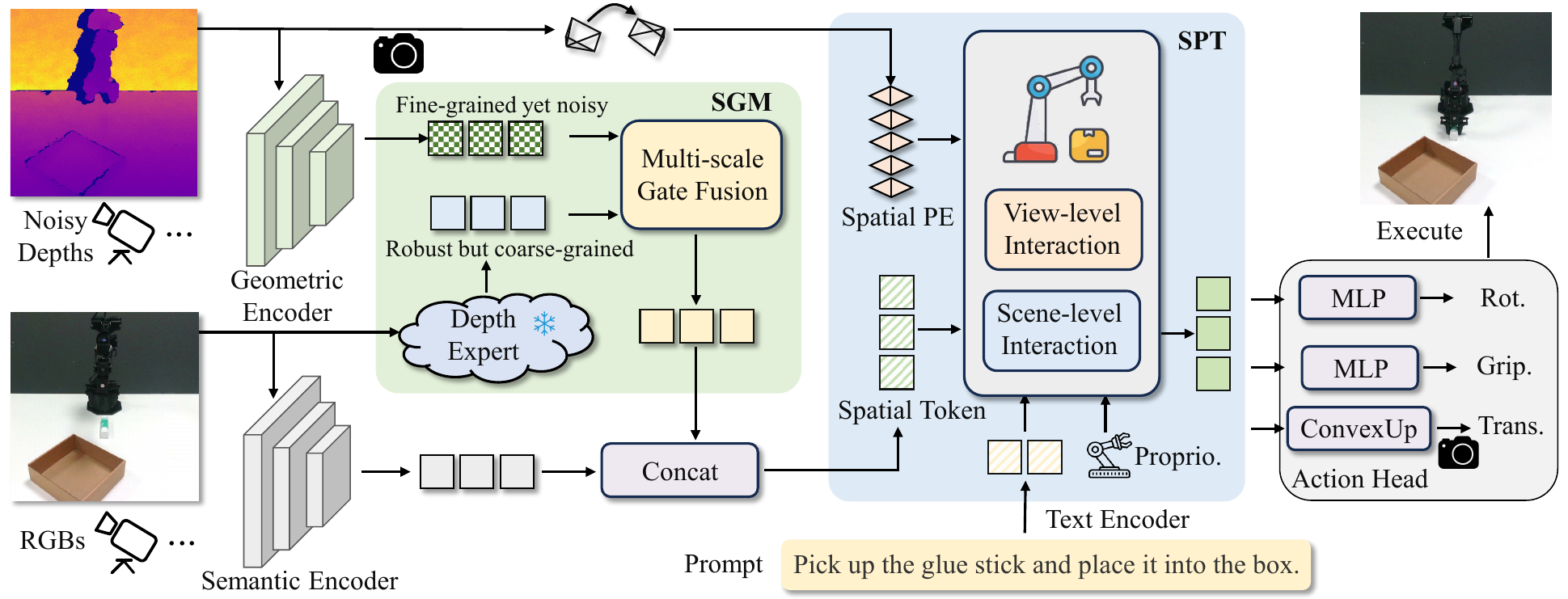}
    \caption{\textbf{Overall framework of~\method.} The architecture employs separate vision and depth encoders. Semantic-guided Geometric Module (SGM) adaptively fuses robust yet coarse geometric priors from a pretrained depth expert with noisy depth features via gated fusion to yield high-level geometric representations. 
    In the Spatial Transformer (SPT), low-level spatial cues are encoded as positional embeddings to drive spatial interactions. 
    Finally, view-level interactions refine intra-view features, while scene-level interactions consolidate cross-modal information across views to support the subsequent action head.}
    \label{fig:overview}
\end{figure*}

\subsection{Representation Learning for Manipulation}
Early methods relied on proprioceptive sensing~\cite{deng2020self,andrychowicz2020learning}, which limited their generalization. 
With the rise of large-scale visual pretraining, many 2D-based approaches~\cite{nair2022r3m,chi2023diffusion,zhao2023learning,yue2025deer,zeng2024learning,zhong2025survey,xie2025dexbotic} leverage strong visual priors to extract semantics. 
However, they often lack 3D spatial understanding, limiting their effectiveness in precise manipulation. 
Point cloud-based methods~\cite{fang2023anygrasp,chen2023polarnet,jia2024lift3d,ze20243d,zhang2023universal,sun2025geovla} capture explicit 3D structures, offering geometry but are hampered by sparsity. 
Voxel-based representations~\cite{shridhar2023perceiver,james2022coarse} reduce sparsity by discretizing space for structured reasoning, yet they incur high computational costs. 
Multi-view RGB-D approaches~\cite{goyal2023rvt,goyal2024rvt,zhang2024sam,fang2025sam2act,wang2024vihe,seo2023multi} integrate dense 2D semantics with geometry via early fusion or auxiliary supervision, yet such shared feature spaces remain vulnerable to sensor noise and often lack precise spatial corresponding for fine-grained interaction. 
To address these limitations, we decouple semantics and geometry, and construct geometric representations by fusing complementary high-level expert priors and raw depth together with low-level spatial cues for precise manipulation. 

\subsection{Vision Foundation Models for Robotics}
Vision foundation models have significantly enhanced robotic perception by incorporating semantic and geometric priors. 
Visual and multimodal models~\cite{radford2021learning,li2022blip,feng2023open,liu2024grounding,wang2025emulating,wu2025ragnet} leverage diverse datasets to learn strong semantic priors that improve visual understanding, which benefits downstream robotic tasks. 
However, they focus on the 2D domain and lack spatial understanding capabilities. 
3D vision models~\cite{zhu2024llava, zheng2024denseg, qian2022pointnext, zheng2025densegrounding, kang2024far, zhang2025grounding} integrate semantic information with explicit spatial structures to facilitate effective geometric perception. 
However, the acquisition and annotation of 3D data are inherently expensive and labor-intensive, which restricts scalability and limits their application in real-world scenarios. 
Depth estimation experts~\cite{yang2024depth,yang2025depth,bhat2023zoedepth,wang2025vggt} leverage large-scale pretraining on diverse datasets to translate semantics in images into corresponding geometric structures, robustly inferring geometric information even under challenging conditions such as sensor noise and occlusions. 
In this paper, we leverage the strong semantic alignment of vision models together with robust geometric priors from depth estimation experts. 

\section{Method}
\label{sec:method}

\begin{figure*}[h]
    \centering
    \includegraphics[width=1.0\linewidth]{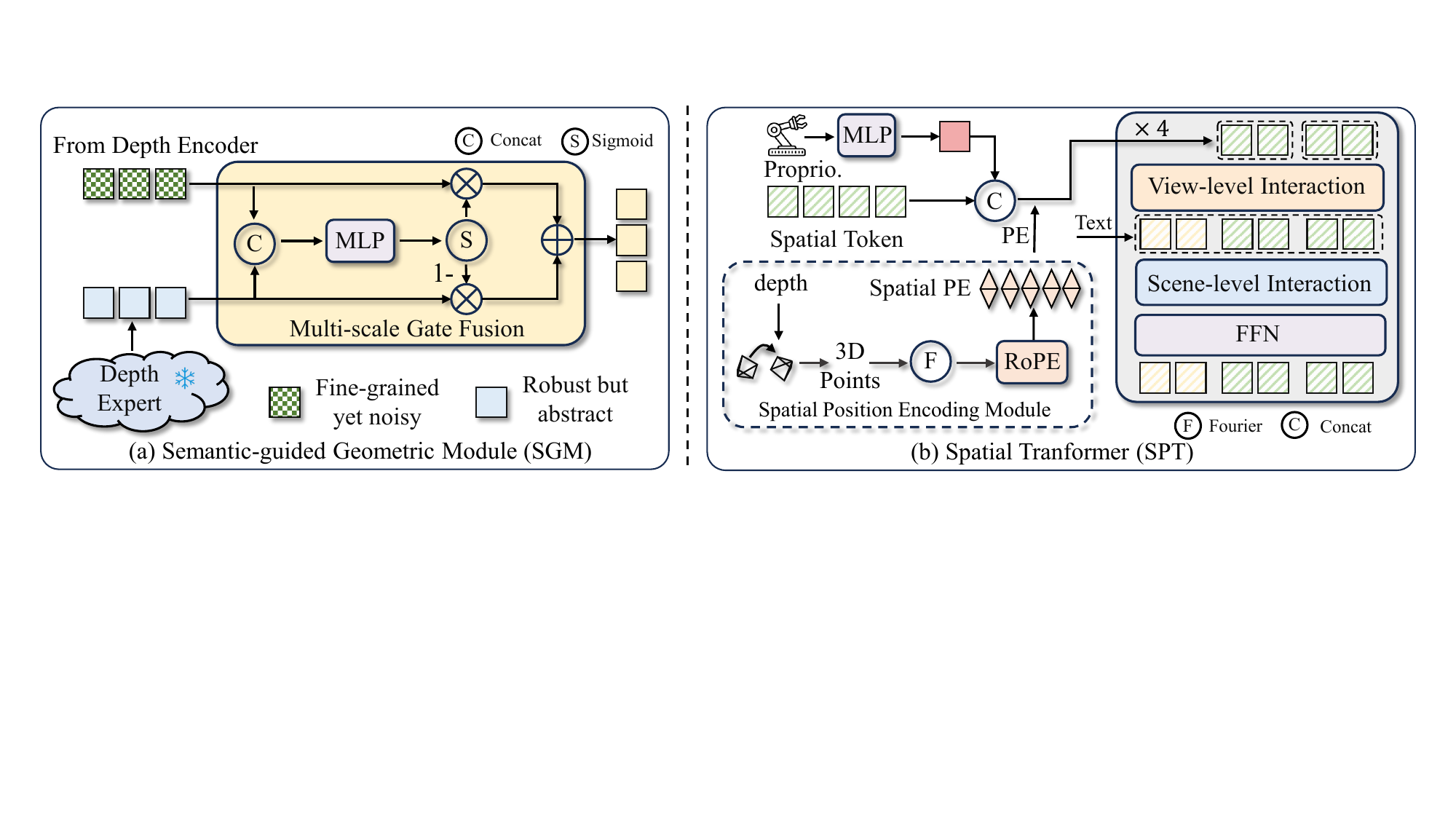}
    \caption{\textbf{Semantic-guided Geometric Module and Spatial Transformer.} 
    (a) SGM adaptively combines two complementary geometric representations via a gating mechanism. 
    (b) SPT converts 3D points into spatial positional embeddings using RoPE to establish 2D–3D correspondences, followed by view-level and scene-level interactions for spatial token refinement. }
    \label{fig:module}
\end{figure*}

\subsection{Overall Framework}
\label{sec:overall}
Fig.~\ref{fig:overview} illustrates the overall framework of our approach. The inputs to the robot's control system are given by
\begin{equation}
    X = \{ I^v, D^v \}_{v=1}^{V},\, P,\, L,
\end{equation}
where \(I^v \in \mathbb{R}^{H \times W \times 3}\) and \(D^v \in \mathbb{R}^{H \times W}\) denote the RGB image and depth map for view \(v\) (with \(V\) views in total), \(P \in \mathbb{R}^{d_p}\) represents the robot’s proprioceptive state (\(d_p\) indicating its dimension), and \(L\) denotes the language prompt. 

For each view \(v\), the RGB images and noisy depth maps are processed separately. 
The images \(I^v\) and the language instruction \(L\) are fed into a vision-language model (e.g., CLIP~\cite{radford2021learning}) to extract semantic features \({F}_{\text{sem}}^{v}\) and text features \(F_{\text{text}}\). 
Meanwhile, raw depth maps \(D^v\) are processed by a depth encoder to yield fine-grained but noisy geometric features \(F_{\text{geo}}^{v}\). 
Subsequently, \(F_{\text{geo}}^{v}\) is enhanced via a Semantic-guided Geometric Module (SGM). In SGM, large-scale pre-trained depth estimation expert is employed to obtain robust yet coarse geometric priors \(\hat{F}_{\text{geo}}^{v}\). A multi-scale gated fusion module then adaptively fuses \(F_{\text{geo}}^{v}\) with \(\hat{F}_{\text{geo}}^{v}\) to produce refined geometric features \(F_{\text{fuse-geo}}^{v}\), preserving details while reducing noise, which are concatenated with \(F_{\text{sem}}^{v}\) to form the final spatial representation \(H^v\). 

We further introduce a Spatial Transformer (SPT). Within the SPT, intrinsic and extrinsic parameters, along with depth values, are used to construct a spatial encoding that captures the low-level spatial cues between spatial tokens. The SPT first applies view-level interaction to consolidate intra-view context, followed by scene-level cross-modal interaction to aggregate cross-modal cues into a unified scene representation. 
Finally, an action head predicts the robot’s 3D end-effector pose and gripper state. 

\subsection{Semantic-guided Geometric Module}
\label{sec:sgm}
Real-world depth measurements are often noisy due to sensor limitations and environmental interference, whereas RGB images provide high signal-to-noise semantic cues. 
Large-scale pretrained depth estimation models (e.g., Depth Anything~\cite{yang2024depth,yang2025depth}) learn a smooth semantic-to-geometric mapping, offering robust and generalizable geometric priors. 
In contrast, raw depth features retain fine-grained, pixel-level details but are highly sensitive to noise. 

To leverage these complementary strengths, we extract robust yet coarse-grained geometric priors from RGB inputs via a frozen large-scale pre-trained depth estimation expert:
\begin{equation}
    \hat{F}_{\text{geo}}^{v} = \mathcal{E}_{\text{expert}}(I^v) \in \mathbb{R}^{H \times W \times C},
\end{equation}
and extract fine-grained but noisy geometry from raw depth using a depth encoder~(e.g. ResNet-50~\cite{he2016deep}): 
\begin{equation}
    F_{\text{geo}}^{v} = \mathcal{E}_{\text{raw}}(D^v) \in \mathbb{R}^{H \times W \times C}.
\end{equation}

As shown in Fig.~\ref{fig:module}~(a), a multi-scale gating mechanism then adaptively fuses these features to yield an optimized geometric representation that preserves fine details while reducing noise and aligning with the semantic cues. 
\begin{equation}
G^{v} = \sigma\bigl(\mathrm{MLP}\bigl(\mathrm{Concat}(\hat{F}_{\text{geo}}^{v},\, F_{\text{geo}}^{v})\bigr)\bigr),
\end{equation}
\begin{equation}
F_{\text{fuse-geo}}^{v} = G^{v} \odot F_{\text{geo}}^{v} + \bigl(1 - G^{v}\bigr) \odot \hat{F}_{\text{geo}}^{v},
\end{equation}
where $ \sigma $ denotes sigmoid activation and $ \odot $ element-wise multiplication. The gate $ G^{v} $ learns to retain reliable depth details while suppressing noise.

\subsection{Spatial Transformer}
\label{sec:spt}
For each view \( v \), we denote the spatial features as \(H^v \in \mathbb{R}^{N_v \times D}\). The proprioceptive input \( P \) is projected via an MLP and fused with \( H^v \) by element-wise addition: 
\begin{equation}
    \widetilde{H}^v = H^v + \text{MLP}(P).
\end{equation}

Given a pixel \( (x', y') \) with depth \( d = D^v(x', y') \), its 3D coordinate \( [x, y, z]^\top \) in the robot-centric coordinate system is computed via perspective projection:
\begin{equation}
[x, y, z, 1]^\top = E^v \left( d \cdot (K^v)^{-1}[x', y', 1]^\top \,\Vert\, 1 \right),
\end{equation}
where \( K^v \in \mathbb{R}^{3 \times 3} \) and \( E^v \in \mathbb{R}^{4 \times 4} \) denote the intrinsic and extrinsic matrices, and \( \Vert \) denotes vector concatenation. 

To encode spatial cues, we apply rotary positional encoding to \( \widetilde{H}^v\), where each axis is assigned \( D/3 \) dimensions. 
We define a set of frequencies: 
\begin{equation}
\omega_k = \lambda^{-2k/d},\quad k = 0, 1, \dots, \frac{d}{2} - 1,\quad d = D/3,
\end{equation}
with \(\lambda=10000\) to control the frequency bandwidth. 
In the spirit of Fourier feature mappings, we compute axis-wise sinusoidal embeddings as: 
\begin{equation}
\cos_{\text{pos}} = \left[\cos(\omega_k u)\right]_{u \in \{x, y, z\},\ k = 0, \dots, d/2 - 1},
\end{equation}
\begin{equation}
\sin_{\text{pos}} = \left[\sin(\omega_k u)\right]_{u \in \{x, y, z\},\ k = 0, \dots, d/2 - 1}.
\end{equation}
The final position-encoded features are given by:
\begin{equation}
T^v = \widetilde{H}^v \odot \cos_{\text{pos}} + \mathrm{rot}(\widetilde{H}^v) \odot \sin_{\text{pos}},
\end{equation}
where \( \odot \) denotes element-wise multiplication, and \( \mathrm{rot}(\cdot) \) rotates each \((f_{2i}, f_{2i+1})\) feature pair as \( (-f_{2i+1}, f_{2i}) \).

At the view level, self-attention followed by a feed-forward network (FFN) refines each view’s token representation. At the scene level, tokens from all views are concatenated with language features \(F_{\text{text}}\). Another round of self-attention and an FFN then fuses cross-view and language context, producing the final refined tokens. 
The tokens are fed into a lightweight decoder (ConvexUp) to generate a per-view 2D heatmap. The target 2D position is obtained via \(\text{argmax}\) and lifted into 3D using the camera model. The action head then uses an MLP on local features around this position to regress the rotation \(\theta = (\theta_x,\theta_y,\theta_z)\) and gripper state \(g\). Together with the 3D translation \((x,y,z)\), these form the final action \(A = (x, y, z, \theta_x, \theta_y, \theta_z, g)\). 

The action supervision includes three parts: a cross-entropy loss on per-view 2D heatmaps for translation, cross-entropy losses on discretized Euler angles for rotation, and a binary classification loss for the gripper state. 

\section{Experiments}
\label{sec:experiments}

\begin{table*}[h]
    \centering
    \resizebox{\textwidth}{!}{
    \begin{tabular}{lcccccccccc}
        \toprule
        Models 
        & \makecell{Avg.\\Success $\uparrow$} 
        & \makecell{Avg.\\Rank $\downarrow$} 
        & \makecell{Close\\Jar} 
        & \makecell{Drag\\Stick} 
        & \makecell{Insert\\Peg} 
        & \makecell{Meat off\\Grill} 
        & \makecell{Open\\Drawer} 
        & \makecell{Place\\Cups} 
        & \makecell{Place\\Wine} 
        & \makecell{Push\\Buttons} \\
        \midrule
C2F-ARM-BC\scriptsize~\cite{james2022coarse} & 20.1 & 9.5 & 24.0 & 24.0 & 4.0 & 20.0 & 20.0 & 0.0 & 8.0  & 72.0 \\
HiveFormer\scriptsize~\cite{guhur2023instruction} & 45.3 & 7.8 & 52.0 & 76.0 & 0.0 & \textbf{100.0} & 52.0 & 0.0 & 80.0 & 84.0 \\
PolarNet\scriptsize~\cite{chen2023polarnet} & 46.4 & 7.3 & 36.0 & 92.0 & 4.0 & \textbf{100.0} & 84.0 & 0.0 & 40.0 & 96.0 \\
PerAct\scriptsize~(Shridhar et al.~\citeyear{shridhar2023perceiver}) & 49.4 & 7.1 & 55.2$_{\pm4.7}$ & 89.6$_{\pm4.1}$ & 5.6$_{\pm4.1}$ & 70.4$_{\pm2.0}$ & 88.0$_{\pm5.7}$ & 2.4$_{\pm3.2}$ & 44.8$_{\pm7.8}$ & 92.8$_{\pm3.0}$ \\
RVT\scriptsize~\cite{goyal2023rvt} & 62.9 & 5.3 & 52.0$_{\pm2.5}$ & 99.2$_{\pm1.6}$ & 11.2$_{\pm3.0}$ & 88.0$_{\pm2.5}$ & 71.2$_{\pm6.9}$ & 4.0$_{\pm2.5}$ & 91.0$_{\pm5.2}$ & \textbf{100.0}$_{\pm0.0}$ \\
Act3D\scriptsize~\cite{gervet2023act3d} & 65.0 & 5.3 & 92.0 & 92.0 & 27.0 & 94.0 & 93.0 & 3.0 & 80.0 & 99.0 \\
SAM-E\scriptsize~\cite{zhang2024sam} & 70.6 & 2.9 & 82.4$_{\pm3.6}$ & \textbf{100.0}$_{\pm0.0}$ & 18.4$_{\pm4.6}$& 95.2$_{\pm3.3}$& \textbf{95.2}$_{\pm5.2}$ & 0.0$_{\pm0.0}$& 94.4$_{\pm4.6}$& \textbf{100.0}$_{\pm0.0}$ \\
3D Diffuser Actor{\scriptsize~(Ke et al.~\citeyear{ke20243d})} & 81.3 & 2.8 & 96.0$_{\pm2.5}$ & \textbf{100.0}$_{\pm0.0}$ & 65.6$_{\pm4.1}$ & 96.8$_{\pm1.6}$ & 89.6$_{\pm4.1}$ & 24.0$_{\pm7.6}$ & 93.6$_{\pm4.8}$ & 98.4$_{\pm2.0}$ \\
RVT-2\scriptsize~\cite{goyal2024rvt} & 81.4 & 2.8 & \textbf{100.0}$_{\pm0.0}$ & 99.0$_{\pm1.7}$ & 40.0$_{\pm0.0}$ & 99.0$_{\pm1.7}$ & 74.0$_{\pm11.8}$ & 38.0$_{\pm4.5}$ & \textbf{95.0}$_{\pm3.3}$ & \textbf{100.0}$_{\pm0.0}$ \\
\midrule
\rowcolor{gray!10}
\method~(Ours)& \textbf{87.4}$_{\pm0.8}$ & \textbf{2.3} & 94.0$_{\pm4.2}$ & \textbf{100.0}$_{\pm0.0}$ & \textbf{93.3}$_{\pm4.8}$ & 98.7$_{\pm2.1}$ & 82.0$_{\pm3.3}$ & \textbf{56.7}$_{\pm8.5}$ & 94.7$_{\pm4.8}$ & \textbf{100.0}$_{\pm0.0}$ \\
        \bottomrule
    \end{tabular}}
    \resizebox{\textwidth}{!}{%
    \begin{tabular}{lcccccccccc}
        \toprule
        Models 
        & \makecell{Put in\\Cupboard} 
        & \makecell{Put in\\Drawer} 
        & \makecell{Put in\\Safe} 
        & \makecell{Screw\\Bulb} 
        & \makecell{Slide\\Block} 
        & \makecell{Sort\\Shape} 
        & \makecell{Stack\\Blocks} 
        & \makecell{Stack\\Cups} 
        & \makecell{Sweep\\Dustpan} 
        & \makecell{Turn\\Tap} \\
        \midrule
C2F-ARM-BC\scriptsize~\cite{james2022coarse} & 0.0 & 4.0 & 12.0 & 8.0 & 16.0 & 8.0 & 0.0 & 0.0 & 0.0 & 68.0 \\
HiveFormer\scriptsize~\cite{guhur2023instruction} & 32.0 & 68.0 & 76.0 & 8.0 & 64.0 & 8.0 & 8.0 & 0.0 & 28.0 & 80.0 \\
PolarNet\scriptsize~\cite{chen2023polarnet} & 12.0 & 32.0 & 84.0 & 44.0 & 56.0 & 12.0 & 4.0 & 8.0 & 52.0 & 80.0 \\
PerAct\scriptsize~(Shridhar et al.~\citeyear{shridhar2023perceiver}) & 28.0$_{\pm4.4}$ & 51.2$_{\pm4.7}$ & 84.0$_{\pm3.6}$ & 17.6$_{\pm2.0}$ & 74.0$_{\pm13.0}$ & 16.8$_{\pm4.7}$ & 26.4$_{\pm3.2}$ & 2.4$_{\pm2.0}$ & 52.0$_{\pm0.0}$ & 88.0$_{\pm4.4}$ \\
RVT\scriptsize~\cite{goyal2023rvt} & 49.6$_{\pm3.2}$ & 88.0$_{\pm5.7}$ & 91.2$_{\pm3.0}$ & 48.0$_{\pm5.7}$ & 81.6$_{\pm5.4}$ & 36.0$_{\pm2.5}$ & 28.8$_{\pm3.9}$ & 26.4$_{\pm8.2}$& 72.0$_{\pm0.0}$ & 93.6$_{\pm4.1}$ \\
Act3D\scriptsize~\cite{gervet2023act3d} & 51.0 & 90.0 & 95.0 & 47.0 & 93.0 & 8.0 & 12.0 & 9.0 & 92.0 & 94.0 \\
SAM-E\scriptsize~\cite{zhang2024sam} & 64.0$_{\pm2.8}$ & 92.0$_{\pm5.7}$ & 95.2$_{\pm3.3}$ & 78.4$_{\pm3.6}$ & 95.2$_{\pm1.8}$ & 34.4$_{\pm6.1}$ & 26.4$_{\pm4.6}$ & 0.0$_{\pm0.0}$& \textbf{100.0}$_{\pm0.0}$ & \textbf{100.0}$_{\pm0.0}$ \\
3D Diffuser Actor{\scriptsize~(Ke et al.~\citeyear{ke20243d})} & \textbf{85.6}$_{\pm4.1}$ & 96.0$_{\pm3.6}$ & \textbf{97.6}$_{\pm2.0}$ & 82.4$_{\pm2.0}$ & \textbf{97.6}$_{\pm3.2}$ & 44.0$_{\pm4.4}$ & 68.3$_{\pm3.3}$ & 47.2$_{\pm8.5}$ & 84.0$_{\pm4.4}$ & 99.2$_{\pm1.6}$ \\
RVT-2\scriptsize~\cite{goyal2024rvt} & 66.0$_{\pm4.5}$ & 96.0$_{\pm0.0}$ & 96.0$_{\pm2.8}$ & 88.0$_{\pm4.9}$ & 92.0$_{\pm2.8}$ & 35.0$_{\pm7.1}$ & \textbf{80.0}$_{\pm2.8}$ & 69.0$_{\pm5.9}$ & \textbf{100.0}$_{\pm0.0}$ & 99.0$_{\pm1.7}$ \\
\midrule
\rowcolor{gray!10}
\method~(Ours)& 72.0$_{\pm3.6}$ & \textbf{98.7}$_{\pm3.3}$ & 96.7$_{\pm3.9}$ & \textbf{88.7}$_{\pm3.9}$ & 91.3$_{\pm6.9}$ & \textbf{73.3}$_{\pm6.5}$ & 56$_{\pm7.6}$ & \textbf{81.3}$_{\pm4.1}$ & \textbf{100.0}$_{\pm0.0}$ & 95.3$_{\pm3.0}$ \\
        \bottomrule
    \end{tabular}}
    \caption{\textbf{Performance on RLBench.} 
    We report success rates on 18 RLBench tasks with 249 variations. 
    \method~achieves the highest overall performance, surpassing the previous state-of-the-art RVT-2 by 6.0\%. 
    Notably, on tasks requiring high spatial precision, such as \texttt{Insert Peg} and \texttt{Sort Shape}, 
    \method~outperforms RVT-2 by 53.3\% and 38.3\%, respectively.}
    \label{tab:rlbench_results}
\end{table*}

To comprehensively evaluate the effectiveness of~\method, we conduct experiments in both simulation and real-world settings. 
Specifically, we aim to answer the following key questions: 
(1) How does~\method~compare to state-of-the-art robotic manipulation policies? 
(2) How robust is \method~under noisy conditions? 
(3) How well does~\method~generalize to few-shot settings? 
(4) How does~\method~perform under spatial perturbations? 
(5) What is the impact of different components of~\method? 
(6) How does~\method~perform in real-robot setups? 

\subsection{Comparison with State-of-the-Art Policies}  
\label{sec:comparison_sota}  
\paragraph{Simulation Environment and Datasets. }
We evaluate \method~on RLBench~\cite{james2020rlbench}, a mainstream multi-task 3D manipulation benchmark built on CoppeliaSim~\cite{rohmer2013v}. 
The simulation environment features a Franka robotic arm with a parallel gripper operating in a tabletop scenario. 
Observations come from four fixed RGB-D cameras (front, left/right shoulder, wrist) at $128 \times 128$ resolution. 
The action space consists of 3D translation, rotation of the end-effector, and binary gripper control. 
An OMPL-based motion planner~\cite{sucan2012open} is utilized to compute feasible trajectories. 
Following PerAct~\cite{shridhar2023perceiver}, we use 18 tasks with 249 variations covering diverse manipulation skills, each with 100 expert demonstrations for training and 25 unseen episodes for evaluation. 

\paragraph{Implementation Details.}
\method~is trained for approximately 40k iterations using a cosine learning rate schedule with an initial 2k-iteration warm-up. 
Training is performed using 8 GPUs with a total batch size of 192 (24 per GPU) and an initial learning rate of $2.4 \times 10^{-3}$. 
Data augmentation includes random spatial translations of up to 12.5 cm along the $x$, $y$, and $z$ axes, as well as rotations of up to $45^\circ$ around the $z$ axis. 
We follow RVT~\cite{goyal2023rvt,goyal2024rvt}, incorporating its virtual view design and two-stage process. 
Furthermore, we employ CLIP~\cite{radford2021learning} as our vision-language encoder to provide aligned cross-modal representations, and Depth Anything v2~\cite{yang2025depth} as our geometry expert.

\paragraph{Performance on RLBench 18 Tasks.} 
Tab.~\ref{tab:rlbench_results} summarizes the performance of various methods on 18 RLBench tasks with 249 variations. 
\method~achieves an average success rate of \textbf{87.4\%}, surpassing the previous state-of-the-art by 6.0\%. 
Notably, \method~shows substantial improvements on tasks requiring high spatial precision, such as \texttt{Insert Peg} and \texttt{Sort Shape}. 
It achieves success rates of 93.3\% and 73.3\% on these tasks, outperforming RVT-2 by \textbf{53.3\%} and \textbf{38.3\%}, respectively. 
These results highlight \method's superior spatial handling capability. 

\subsection{Robustness under Noisy Conditions}  
\label{sec:robustness_noisy}  

\begin{table*}[h]
    \centering
    \resizebox{\textwidth}{!}{
    \begin{tabular}{lcccccccccc}
        \toprule
        Models & Noise type & Avg. Success $\uparrow$ 
               & Close Jar & Drag Stick & Insert Peg & Meat off Grill 
               & Open Drawer & Place Cups & Place Wine & Push Buttons \\
        \midrule
RVT-2
& \multirow{2}{*}{\textit{Light}} & 72.5$_{\pm0.5}$
& 92.0$_{\pm4.0}$ 
& \textbf{100.0}$_{\pm0.0}$
& 6.7$_{\pm4.6}$
& \textbf{100.0}$_{\pm0.0}$
& \textbf{82.7}$_{\pm10.1}$
& 25.3$_{\pm6.1}$
& \textbf{96.0}$_{\pm4.0}$
& 74.7$_{\pm8.3}$ 
\\[3pt]
\method~(Ours)
& & \textbf{86.4}$_{\pm0.4}$
& \textbf{97.3}$_{\pm2.3}$
& 98.7$_{\pm2.3}$
& \textbf{94.7}$_{\pm6.1}$
& 96.0$_{\pm0.0}$
& 73.3$_{\pm10.1}$
& \textbf{54.7}$_{\pm8.3}$
& 92.0$_{\pm4.0}$
& \textbf{98.7}$_{\pm2.3}$
\\
        \midrule
RVT-2
& \multirow{2}{*}{\textit{Middle}} & 68.4$_{\pm0.9}$
& 85.3$_{\pm2.3}$
& \textbf{100.0}$_{\pm0.0}$
& 2.7$_{\pm2.3}$
& 94.7$_{\pm2.3}$
& \textbf{82.7}$_{\pm11.5}$
& 20.0$_{\pm0.0}$
& \textbf{89.3}$_{\pm4.6}$
& 73.3$_{\pm4.6}$
\\[3pt]
\method~(Ours)
& & \textbf{85.3}$_{\pm0.9}$
& \textbf{100.0}$_{\pm0.0}$
& 98.7$_{\pm2.3}$
& \textbf{81.3}$_{\pm6.1}$
& \textbf{96.0}$_{\pm4.0}$
& 78.7$_{\pm8.3}$
& \textbf{45.3}$_{\pm10.1}$
& \textbf{89.3}$_{\pm4.6}$
& \textbf{97.3}$_{\pm4.6}$
\\
        \midrule
RVT-2
& \multirow{2}{*}{\textit{Heavy}} & 57.0$_{\pm0.9}$
& 49.3$_{\pm6.1}$
& 94.7$_{\pm4.6}$
& 0.0$_{\pm0.0}$
& 97.3$_{\pm2.3}$
& \textbf{86.7}$_{\pm2.3}$
& 8.0$_{\pm4.0}$
& 86.7$_{\pm2.3}$
& 64.0$_{\pm4.0}$
\\[3pt]
\method~(Ours)
& & \textbf{76.4}$_{\pm0.5}$
& \textbf{82.7}$_{\pm2.3}$
& \textbf{98.7}$_{\pm2.3}$
& \textbf{61.3}$_{\pm6.1}$
& \textbf{100.0}$_{\pm0.0}$
& 80.0$_{\pm4.0}$
& \textbf{21.3}$_{\pm4.6}$
& \textbf{92.0}$_{\pm0.0}$
& \textbf{92.0}$_{\pm4.6}$
\\
        \bottomrule
    \end{tabular}}
    \resizebox{\textwidth}{!}{%
    \begin{tabular}{lcccccccccc}
        \toprule
        Models 
        & Put in Cupboard & Put in Drawer & Put in Safe & Screw Bulb 
        & Slide Block & Sort Shape & Stack Blocks & Stack Cups 
        & Sweep to Dustpan & Turn Tap \\
        \midrule
RVT-2
& 57.3$_{\pm2.3}$
& \textbf{100.0}$_{\pm0.0}$
& 92.0$_{\pm4.0}$
& 81.3$_{\pm6.1}$
& 62.7$_{\pm23.1}$
& 46.7$_{\pm6.1}$
& 53.3$_{\pm2.3}$
& 45.3$_{\pm2.3}$
& 96.0$_{\pm6.9}$
& \textbf{93.3}$_{\pm4.6}$
\\[3pt]
\method~(Ours)
& \textbf{81.3}$_{\pm2.3}$
& 98.7$_{\pm2.3}$
& \textbf{98.7}$_{\pm2.3}$
& \textbf{88.0}$_{\pm4.0}$
& \textbf{72.0}$_{\pm4.0}$
& \textbf{76.0}$_{\pm6.9}$
& \textbf{62.7}$_{\pm2.3}$
& \textbf{82.7}$_{\pm2.3}$
& \textbf{97.3}$_{\pm4.6}$
& \textbf{93.3}$_{\pm2.3}$
\\
        \midrule
RVT-2
& 50.7$_{\pm12.2}$
& 98.7$_{\pm2.3}$
& \textbf{98.7}$_{\pm2.3}$
& 76.0$_{\pm4.0}$
& 57.3$_{\pm2.3}$
& 38.7$_{\pm10.1}$
& 45.3$_{\pm12.2}$
& 25.3$_{\pm6.1}$
& 96.0$_{\pm4.0}$
& 96.0$_{\pm6.9}$
\\[3pt]
\method~(Ours)
& \textbf{74.7}$_{\pm6.1}$
& \textbf{100.0}$_{\pm0.0}$
& 94.7$_{\pm2.3}$
& \textbf{88.0}$_{\pm4.0}$
& \textbf{81.3}$_{\pm15.1}$
& \textbf{76.0}$_{\pm4.0}$
& \textbf{58.7}$_{\pm6.1}$
& \textbf{77.3}$_{\pm8.3}$
& \textbf{100.0}$_{\pm0.0}$
& \textbf{97.3}$_{\pm2.3}$
\\
        \midrule
RVT-2
& 20.0$_{\pm6.9}$
& 97.3$_{\pm2.3}$
& 93.3$_{\pm2.3}$
& 58.7$_{\pm2.3}$
& 57.3$_{\pm8.3}$
& 13.3$_{\pm6.1}$
& 13.3$_{\pm6.1}$
& 1.3$_{\pm2.3}$
& \textbf{92.0}$_{\pm0.0}$
& 92.0$_{\pm4.0}$
\\[3pt]
\method~(Ours)
& \textbf{64.0}$_{\pm4.0}$
& \textbf{100.0}$_{\pm0.0}$
& \textbf{100.0}$_{\pm0.0}$
& \textbf{78.7}$_{\pm8.3}$
& \textbf{58.7}$_{\pm2.3}$
& \textbf{52.0}$_{\pm4.0}$
& \textbf{42.7}$_{\pm6.1}$
& \textbf{70.7}$_{\pm6.1}$
& 82.7$_{\pm2.3}$
& \textbf{97.3}$_{\pm4.6}$
\\
        \bottomrule
    \end{tabular}}
    \caption{\textbf{Performance Under Various Noise Levels.} We report success rates under three noise conditions: \textit{Light} noise corrupts 20\% of the points in the reconstructed point cloud with random Gaussian noise (std = 0.05), \textit{Middle} noise corrupts 50\% with noise of std = 0.1, and \textit{Heavy} noise corrupts 80\% with noise of std = 0.1. Under these conditions, \method~improves average success rates by approximately 13.9\%, 16.9\%, and 19.4\% over RVT-2 at the Light, Middle, and Heavy noise levels, respectively. }
    \label{tab:noise_results}
\end{table*}

\paragraph{Experimental Setup.}  
Depth measurements are inherently affected by sensor noise, lighting variations, and surface reflections. To simulate these challenges, we inject controlled Gaussian noise into reconstructed point clouds. 
Specifically, we design three noise levels: \emph{Light} corrupts 20\% of the points with a Gaussian standard deviation of 0.05, \emph{Middle} corrupts 50\% of the points with a standard deviation of 0.1, and \emph{Heavy} corrupts 80\% of the points with a standard deviation of 0.1. 
This setup allows us to evaluate the robustness of our approach under progressively severe noisy conditions. 

\paragraph{Performance Evaluation.} 
Tab.~\ref{tab:noise_results} shows that under \textit{Light}, \textit{Middle}, and \textit{Heavy} noise, \method~improves average success rates over RVT-2 by 13.9\%, 16.9\%, and 19.4\%, respectively. 
Notably, in tasks requiring high spatial precision, these gains are even more pronounced. 
For instance, on \texttt{Insert Peg} task, \method~outperforms RVT-2 by 88.0\%, 78.6\%, and 61.3\% under the respective noise levels. 

\label{sec:few_shot}  
\begin{table*}[t]
\centering
\resizebox{0.85\textwidth}{!}{%
\begin{tabular}{lcccccccccc}
\toprule
Models 
& \makecell{Avg.\\Success $\uparrow$} 
& \makecell{Close\\Laptop} 
& \makecell{Put Rubbish\\in Bin} 
& \makecell{Beat\\Buzz} 
& \makecell{Close\\Microwave} 
& \makecell{Put Shoes\\in Box} 
& \makecell{Get\\Ice} 
& \makecell{Change\\Clock} 
& \makecell{Close\\Box} 
& \makecell{Reach\\Target} \\
\midrule
RVT-2
& 46.9\textsubscript{\(\pm1.5\)} 
& 76.0\textsubscript{\(\pm6.1\)} 
& 10.3\textsubscript{\(\pm5.1\)} 
& 47.4\textsubscript{\(\pm8.5\)} 
& 61.7\textsubscript{\(\pm9.8\)} 
& 7.4\textsubscript{\(\pm4.3\)} 
& 93.7\textsubscript{\(\pm3.9\)} 
& 72.6\textsubscript{\(\pm2.8\)} 
& 49.1\textsubscript{\(\pm8.6\)} 
& 12.0\textsubscript{\(\pm5.7\)} 
\\
\method
& \textbf{79.2}\textsubscript{\(\pm2.7\)}
& \textbf{90.0}\textsubscript{\(\pm7.5\)}
& \textbf{100.0}\textsubscript{\(\pm0.0\)}
& \textbf{92.0}\textsubscript{\(\pm2.5\)}
& \textbf{95.3}\textsubscript{\(\pm11.4\)}
& \textbf{25.3}\textsubscript{\(\pm13.8\)}
& \textbf{96.0}\textsubscript{\(\pm2.5\)}
& \textbf{83.3}\textsubscript{\(\pm7.3\)}
& \textbf{95.3}\textsubscript{\(\pm4.7\)}
& \textbf{86.0}\textsubscript{\(\pm2.2\)}
\\
\bottomrule
\end{tabular}}
\resizebox{0.85\textwidth}{!}{%
\begin{tabular}{lcccccccccc}
\toprule
Models 
& \makecell{Close\\Door} 
& \makecell{Remove\\Cups} 
& \makecell{Close\\Drawer} 
& \makecell{Spatula\\Scoop} 
& \makecell{Close\\Fridge} 
& \makecell{Put Knife\\on Board} 
& \makecell{Screw\\Nail} 
& \makecell{Close\\Grill} 
& \makecell{Plate\\in Rack} 
& \makecell{Meat on\\Grill} \\
\midrule
RVT-2
& 4.0\textsubscript{\(\pm3.3\)}
& 33.7\textsubscript{\(\pm13.8\)}
& 96.0\textsubscript{\(\pm0.0\)}
& 70.9\textsubscript{\(\pm6.8\)}
& 81.7\textsubscript{\(\pm8.6\)}
& 14.3\textsubscript{\(\pm7.3\)}
& 38.9\textsubscript{\(\pm15.1\)}
& 66.3\textsubscript{\(\pm8.9\)}
& 24.6\textsubscript{\(\pm7.1\)}
& 30.0\textsubscript{\(\pm8.5\)}
\\
\method
& \textbf{36.0}\textsubscript{\(\pm14.1\)}
& \textbf{66.0}\textsubscript{\(\pm8.3\)}
& \textbf{96.7}\textsubscript{\(\pm3.9\)}
& \textbf{84.7}\textsubscript{\(\pm8.2\)}
& \textbf{95.3}\textsubscript{\(\pm5.3\)}
& \textbf{66.0}\textsubscript{\(\pm2.2\)}
& \textbf{62.7}\textsubscript{\(\pm6.0\)}
& \textbf{96.0}\textsubscript{\(\pm0.0\)}
& \textbf{48.0}\textsubscript{\(\pm8.0\)}
& \textbf{90.0}\textsubscript{\(\pm2.8\)}
\\
\bottomrule
\end{tabular}}
\caption{\textbf{Few-Shot Generalization.} We adapt pre-trained model to 19 new tasks using only 10 demonstrations per task (1/10th of original data). We reports success rates, showing that \method, significantly outperforms RVT-2 in the few-shot setting. }
\label{tab:few_shot_results}
\end{table*}

\subsection{Few-Shot Generalization}  
We evaluate the few-shot generalization ability of \method~by adapting the multi-task pre-trained model to 19 novel tasks using only 10 demonstrations per task, just one-tenth of the data used during multi-task training. 
In this few-shot adaptation scenario, the model is initialized with its pre-trained weights and then fine-tuned on the limited data. 
As shown in Tab.~\ref{tab:few_shot_results}, our experiments demonstrate that \method~effectively transfers previously learned skills to new tasks with minimal adaptation data. Overall, \method~achieves an average success rate of 79.2\%, compared to 46.9\% for RVT-2, yielding an improvement of approximately 32.3\%. This significant boost underscores the superior few-shot generalization capability of our approach. 

\subsection{Spatial Perturbations on ColosseumBench} 
\label{sec:colosseum_results}

\paragraph{Setup.}  
We evaluate the robustness on the Colosseum benchmark~\cite{pumacay2024colosseum}, which assesses manipulation policies under environmental variations. 
We report results on 20 tasks under the baseline setting (no perturbation) and three spatial perturbations: manipulation object size (MO-Size), which scales the object being manipulated; receiver object size (RO-Size), which scales an indirectly used object such as a container; and camera pose perturbation (Camera-Perturb), which randomly shifts the camera’s position and orientation to vary the observation viewpoint.

\paragraph{Performance Evaluation.}  
The results in Tab.~\ref{tab:colosseum_results} indicate that under the no-perturbation condition, our method achieves a task-average success rate of 57.4\%. 
When spatial perturbations are introduced, \method~attains 59.2\% under MO-Size variations, 62.0\% under RO-Size changes, and 54.2\% with camera pose perturbations. 
These results consistently outperform competing methods, demonstrating strong robustness and generalization under spatial variations. 

\subsection{Ablation Study}  
\label{sec:ablation}  

Our ablation study on 18 tasks (Tab.~\ref{tab:ablation}) shows that decoupling semantics and geometry improves performance in both noise-free and heavy-noise settings, increasing success rates from 81.4\% to 85.1\% and from 57.0\% to 68.7\%, respectively. 
Introducing the Semantic-guided Geometry Module (SGM) further boosts performance, especially under heavy noise, where performance rises to 73.9\%. 
Finally, the Spatial Transformer (SPT), which provides precise low-level spatial cues, brings the success rates to 87.4\% and 76.4\% in noise-free and noisy conditions, respectively. 

\begin{table}[t]
\centering
\resizebox{0.46\textwidth}{!}{
\begin{tabular}{lcccc}
\toprule
\textbf{Method}
& \makecell{\textbf{No-}\\\textbf{Vars} $\uparrow$}
& \makecell{\textbf{MO-}\\\textbf{Size} $\uparrow$}
& \makecell{\textbf{RO-}\\\textbf{Size} $\uparrow$}
& \makecell{\textbf{Cam}\\\textbf{Pose} $\uparrow$} \\
\midrule
R3M{\scriptsize~\cite{nair2022r3m}} & 2.9 & 1.8 & 0.0 & 0.8 \\
MVP{\scriptsize~\cite{Radosavovic2022}} & 3.4 & 4.4 & 0.5 & 2.6 \\
VoxPoser{\scriptsize~\cite{huang2023voxposer}} & 5.4 & 3.3& 6.5 & 6.2 \\
PerAct{\scriptsize~(Shridhar et al.~\citeyear{shridhar2023perceiver})} & 34.5 & 35.6 & 29.3 & 36.3 \\
RVT{\scriptsize~\cite{goyal2023rvt}} & 43.6 & 35.3 & 40.5 & 42.2 \\
\midrule
\rowcolor{gray!10}
\method~(Ours) & \textbf{57.4}$_{\pm3.0}$ & \textbf{59.2}$_{\pm2.4}$ & \textbf{62.0}$_{\pm3.2}$ & \textbf{54.2}$_{\pm1.8}$ \\
\toprule
\end{tabular}
}
\caption{\textbf{Performance Under Spatial Perturbations.} 
We report average success rates on 20 ColosseumBench tasks under 4 conditions: No-Vars, manipulation object size, receiver object size, and camera pose.}
\label{tab:colosseum_results}
\end{table}

\begin{figure}[h]
    \centering
    \includegraphics[width=0.925\linewidth]{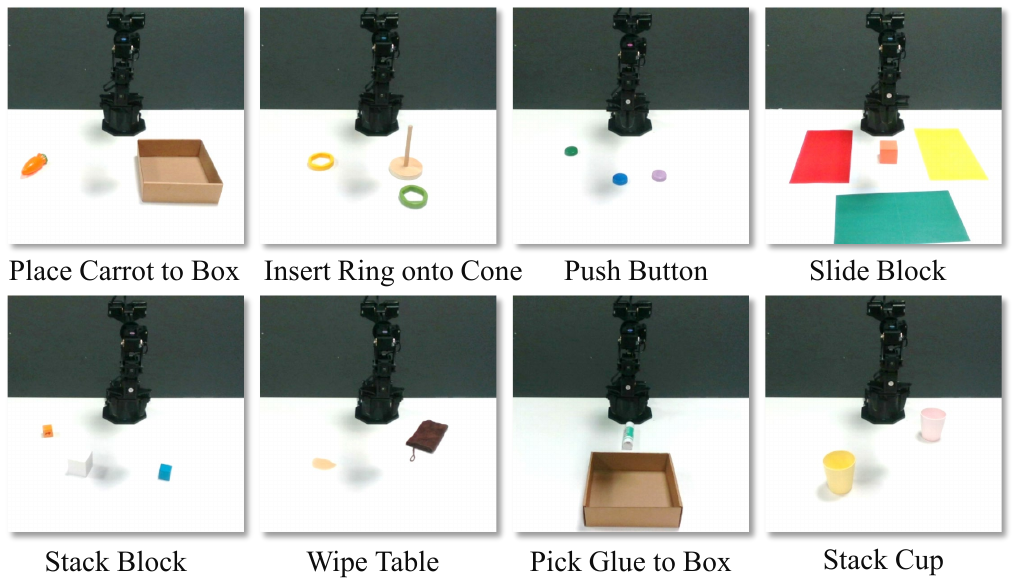}
    \caption{\textbf{Real-world tasks.} We employed 8 distinct tasks with a total of 15 variants in real-world experiments. }
    \label{fig:real_tasks}
\end{figure}

\subsection{Real-World Evaluation}  
\label{sec:real_world}  

\paragraph{Setup.} 
In real-world experiments, we use a WidowX single-arm robot equipped with an Intel RealSense D435i RGB-D camera. The camera is statically mounted to capture a front view of the workspace. We perform both intrinsic and extrinsic calibration between the camera and the robot to accurately transform the observed point clouds into the robot's base coordinate system. The system is integrated using a ROS package. Images are originally captured at a resolution of $1280 \times 720$ and are downsampled to $128 \times 128$. 

\begin{table}[t]
    \centering
    \begin{tabular}{ccccc}
        \toprule
        \textbf{Decouple}  &\textbf{SGM}& \textbf{SPT} & \multicolumn{2}{c}{\textbf{Avg. success on 18 tasks} $\uparrow$} \\
        \cmidrule(lr){4-5}
          &&  & \textbf{No noise} & \textbf{Heavy noise} \\
        \midrule
          &&  & 81.4 & 57.0 \\
        \checkmark  &&  & 85.1 &  68.7\\
        \checkmark  &\checkmark &  & 86.4 &  73.9\\
        \checkmark  &\checkmark & \checkmark & 87.4 & 76.4 \\
        \bottomrule
    \end{tabular}
    \caption{\textbf{Ablation Study.} We analyze the contribution of each module to overall performance and their effect on robustness under heavy noisy conditions.}
    \label{tab:ablation}
\end{table}

\begin{figure}[t]
    \centering
    \includegraphics[width=1.0\linewidth]{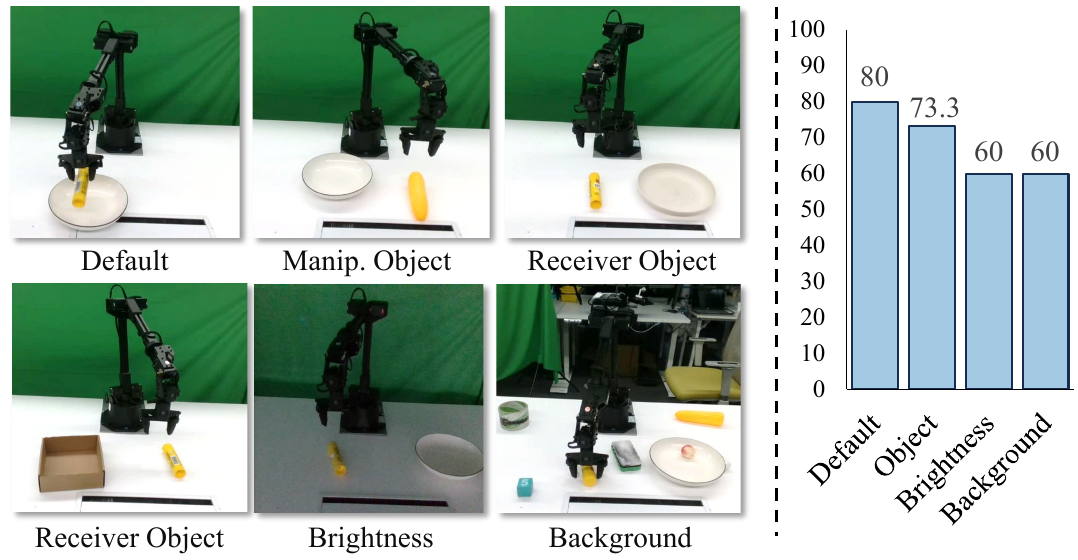}
    \caption{\textbf{Real-world Generalization Evaluation}. 
    We assess~\method~under variations in manipulated object, receiver object, brightness, and background. Performance remains robust across challenging settings.}
    \label{fig:real_ood}
\end{figure}

\begin{table}[h!]
\centering
\resizebox{0.46\textwidth}{!}{%
\begin{tabular}{lcccc}
\toprule
\textbf{Task} & \textbf{\#Var.} & \textbf{RVT-2} & \textbf{\method} \\
\midrule
(1) Pick Glue to Box & 1 & 50\% & 85\% \\
(2) Stack Cup & 2 & 30\% & 30\% \\
(3) Push Button & 3 & 67\% & 90\% \\
(4) Slide Block & 3 & 60\% & 67\% \\
(5) Place Carrot to Box & 1 & 30\% & 65\% \\
(6) Stack Block & 2 & 40\% & 35\% \\
(7) Insert Ring Onto Cone & 2 & 20\%  & 50\% \\
(8) Wipe Table & 1 & 50\% & 80\% \\
All tasks & 15 & 43\% & 63\% \\
\bottomrule
\end{tabular}
}
\caption{\textbf{Real-World Results.} We report success rates for each task and overall performance across 8 tasks with 15 variations. \method, consistently outperforms RVT-2, indicating superior robustness in real-world scenarios.}
\label{tab:real_world_results}
\end{table}

\paragraph{Dataset Collection.}  
We conduct experiments on a series of real-world tasks (Fig.~\ref{fig:real_tasks}), including (1) Pick Glue to Box, (2) Stack Cup, (3) Push Button, (4) Slide Block, (5) Place Carrot to Box, (6) Stack Block, (7) Insert Ring onto Cone, and (8) Wipe Table. For each task, we collect 25 demonstrations that capture diverse spatial configurations and object variations. Some tasks are instantiated with multiple variations, for example, the Slide Block task includes yellow, green, and red variants, resulting in a total of 15 variations across the 8 tasks. The trajectories are recorded at 30 fps, and key-frames are extracted to construct the training set. 

\paragraph{Evaluation.}  
We evaluate~\method~against RVT-2 on various real-world tasks. 
Single-variant tasks are tested 20 times, and multi-variant tasks 10 times per variant. 
As shown in Tab.~\ref{tab:real_world_results}, \method\ consistently outperforms RVT-2, with an average improvement of around 20\% across tasks, demonstrating effectiveness in real-world scenarios. 

To evaluate robustness to distribution shifts, we test~\method~under variations in manipulated object, receiver object, lighting, and background (Fig.~\ref{fig:real_ood}). 
\method~maintains consistently high performance across these diverse and challenging conditions, clearly demonstrating strong robustness and generalization in complex real-world scenarios.

\section{Conclusion}
\label{sec:conclusion}

In this work, we present~\method, a framework for robust spatial representation in robotic manipulation. \method~disentangles semantic and geometric information, with the geometric branch comprising a high-level module (SGM), which fuses semantic-guided geometric priors with depth features, and a low-level module (SPT), which captures fine-grained spatial cues through position-aware interactions. Experiments across 50+ simulated and real-world tasks show that \method~achieves higher success rates and strong robustness, underscoring the value of disentangled spatial representations for reliable manipulation.

\clearpage
\section*{Acknowledgments}
This work was supported by the National Science and Technology Major Project of China under Grant No. 2023ZD0121300, the Scientific Research Innovation Capability Support Project for Young Faculty under Grant No. ZYGXQNJSKYCXNLZCXM-I20, and the National Natural Science Foundation of China under Grant No. U24B20173.

\bibliography{aaai2026}

\clearpage
\appendix
\twocolumn[
\begin{center}
    \LARGE \bf Supplementary Material
\end{center}
\vspace{1em}
]
\vspace{1em}

\section{Hyperparameters}

As detailed in Table \ref{tab:hyperparameters}, we train our model with a per-GPU batch size of 24 on eight GPUs for a total batch size of 192, using the LAMB optimizer at an initial learning rate of $4\times10^{-3}$ and a cosine decay schedule. A linear warmup runs for the first 2000 steps, followed by around 40 000 training steps (50 epochs). 

\section{Data Scaling Analysis}

We examine how the size of the training dataset affects our model’s performance on eight challenging RLBench manipulation tasks. 
As shown in Table~\ref{tab:data_scaling}, the average success rate grows monotonically with the number of training samples, rising from 69.3 \% at 25 samples to 73.0 \% at 50 samples, 75.4 \% at 100 samples, and 80.0 \% at 200 samples, before reaching 81.3 \% at 500 samples. 
This monotonic improvement illustrates a clear data-scaling effect, where enlarging the dataset consistently enhances task performance and generalization. 
For fair comparison with prior work, however, we report all results in the main paper using the 100 sample. 

\section{Ablation on View Setup}

Table~\ref{tab:view} presents success rates for the front-only (single-view) and multi-view configurations on 18 RLBench manipulation tasks. The multi-view setup, which integrates front, left shoulder, and right shoulder cameras, increases the average success rate from 80.0 \% to 87.4 \%, demonstrating that additional viewpoints enrich spatial context and substantially improve manipulation performance. 

\section{Ablation on Visual Backbone}

Table \ref{tab:backbone} compares DINO and CLIP as visual backbones on 18 RLBench manipulation tasks. Using CLIP’s language-aligned visual features increases the average success rate from 86.5 \% to 87.4 \%, indicating that semantic alignment in the encoder contributes to improved task performance. 

\section{Qualitative Results}

\begin{figure*}[h!]
    \centering
    \includegraphics[width=0.7\linewidth]{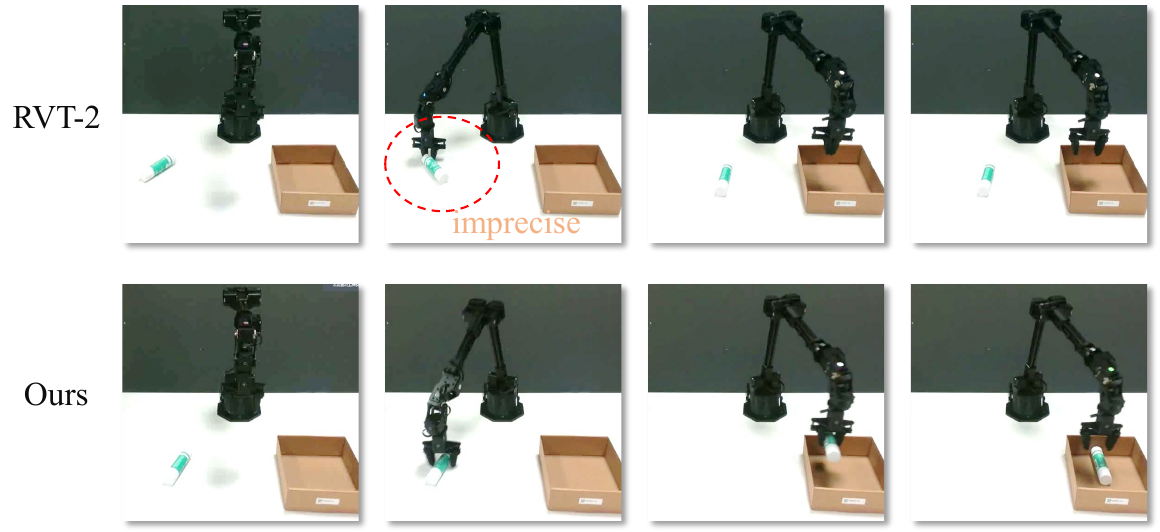}
    \caption{\textbf{Qualitative Comparison on the PlaceGlueToBox Task.} RVT-2 often fails to grasp the glue stick reliably, with its gripper missing or slipping off the object due to noisy depth. In contrast, our method consistently secures the stick and holds it firmly for downstream placement.}
    \label{fig:qualitative_glue}
\end{figure*}

\begin{figure*}[h!]
    \centering
    \includegraphics[width=0.7\linewidth]{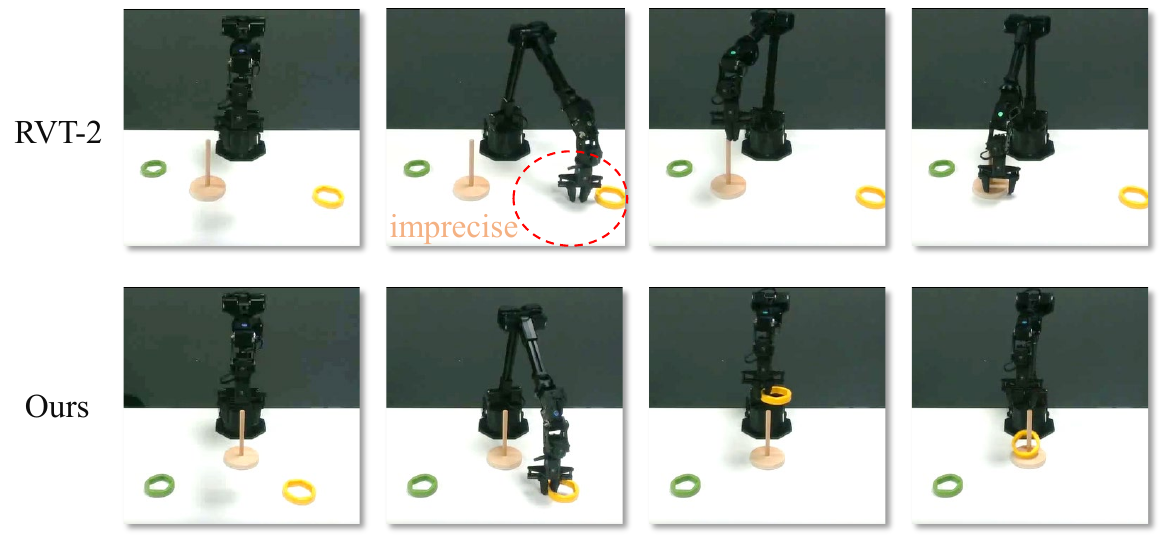}
    \caption{\textbf{Qualitative Comparison on the InsertRingOntoCone Task.} RVT-2’s noisy perception leads to unstable grasps that drop or misalign the ring during pickup. Our method achieves stable grasping and precise alignment of the ring before insertion.}
    \label{fig:qualitative_ring}
\end{figure*}

Figures \ref{fig:qualitative_glue} and \ref{fig:qualitative_ring} offer comparisons between RVT-2 and our method on two real-world manipulation tasks. 
In the PlaceGlueToBox task, RVT-2’s grasp attempts frequently slip or miss the glue stick as a result of noisy depth perception, whereas our method produces stable, secure grasps that reliably lift the stick for subsequent placement. 
In the InsertRingOntoCone task, RVT-2’s unstable pickups often let the ring slip or drift off-axis, whereas our method reliably acquires the ring and centers it above the cone for smooth insertion. 

\section{Failure Cases}

\begin{figure*}[h!]
    \centering
    \includegraphics[width=1\linewidth]{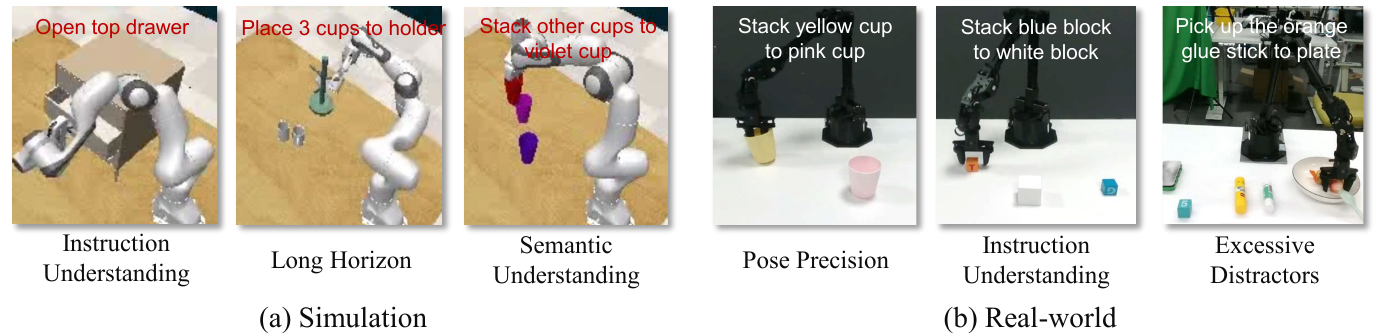}
    \caption{\textbf{Examples of Failure Cases in Simulation and Real-world}. (a) Simulation failures from instruction mis-understanding, long-horizon errors, and semantic understanding mistakes. (b) Real-world failures due to pose precision limits, instruction understanding mistakes, and excessive visual distractors. }
    \label{fig:fail_cases}
\end{figure*}

As shown in Figure~\ref{fig:fail_cases} (a), simulation failures include (1) instruction understanding errors, where the Open Drawer task opens the wrong drawer; (2) long-horizon breakdowns, where Place Cups stalls after only a few placements; and (3) semantic grounding confusion, where Stack Cups picks the incorrect cup among similar ones. 
In real-world trials (Figure~\ref{fig:fail_cases} (b)), failures arise from (1) pose precision limits, where slight end-effector drift causes Stack Cups to miss its target; (2) instruction mis-understanding, where Stack Blocks grasps the wrong block; and (3) distractor susceptibility, where background clutter diverts the policy during Pick Glue To Plate. 
These findings point toward promising enhancements, such as incorporating large language models (LLMs) for more accurate instruction parsing, adding episodic memory or belief tracking to support reliable long-horizon planning, and integrating uncertainty-aware pose refinement and attention-based filtering to improve resilience against calibration errors and visual clutter. 

\section{Robot Setup}

\begin{figure*}[h!]
    \centering
    \includegraphics[width=0.6\linewidth]{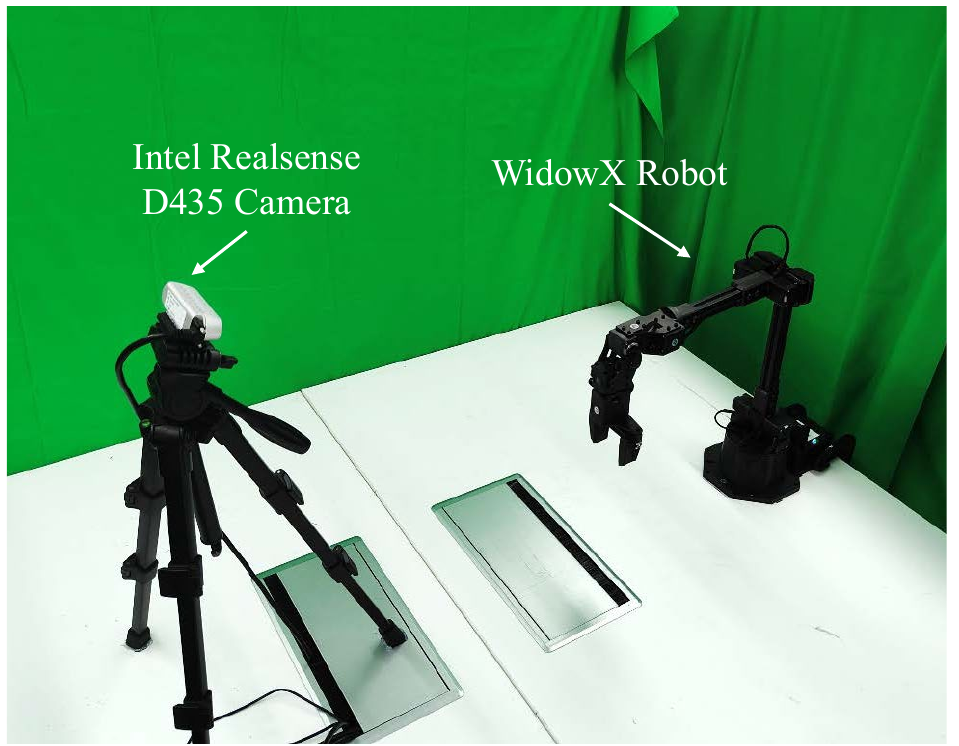}
    \caption{\textbf{Real-World Robot Setup}. WidowX-250 arm and Intel RealSense D435i camera mounted 0.8 m apart in a front-facing configuration.}
    \label{fig:robot_setup}
\end{figure*}

As shown in Figure \ref{fig:robot_setup}, a WidowX-250 single-arm robot sits adjacent to an Intel RealSense D435i RGB-D camera fixed on a tripod 0.8 m from the workspace. The camera captures synchronized 1280×720 color and depth frames at 30 Hz, which are downsampled to 128×128 for input into~\method. We perform intrinsic and extrinsic calibration to align depth measurements with the robot’s base frame, enabling accurate end-effector control in real-world. 

\section{Simulation Tasks}

\begin{figure*}[h!]
    \centering
    \includegraphics[width=1\linewidth]{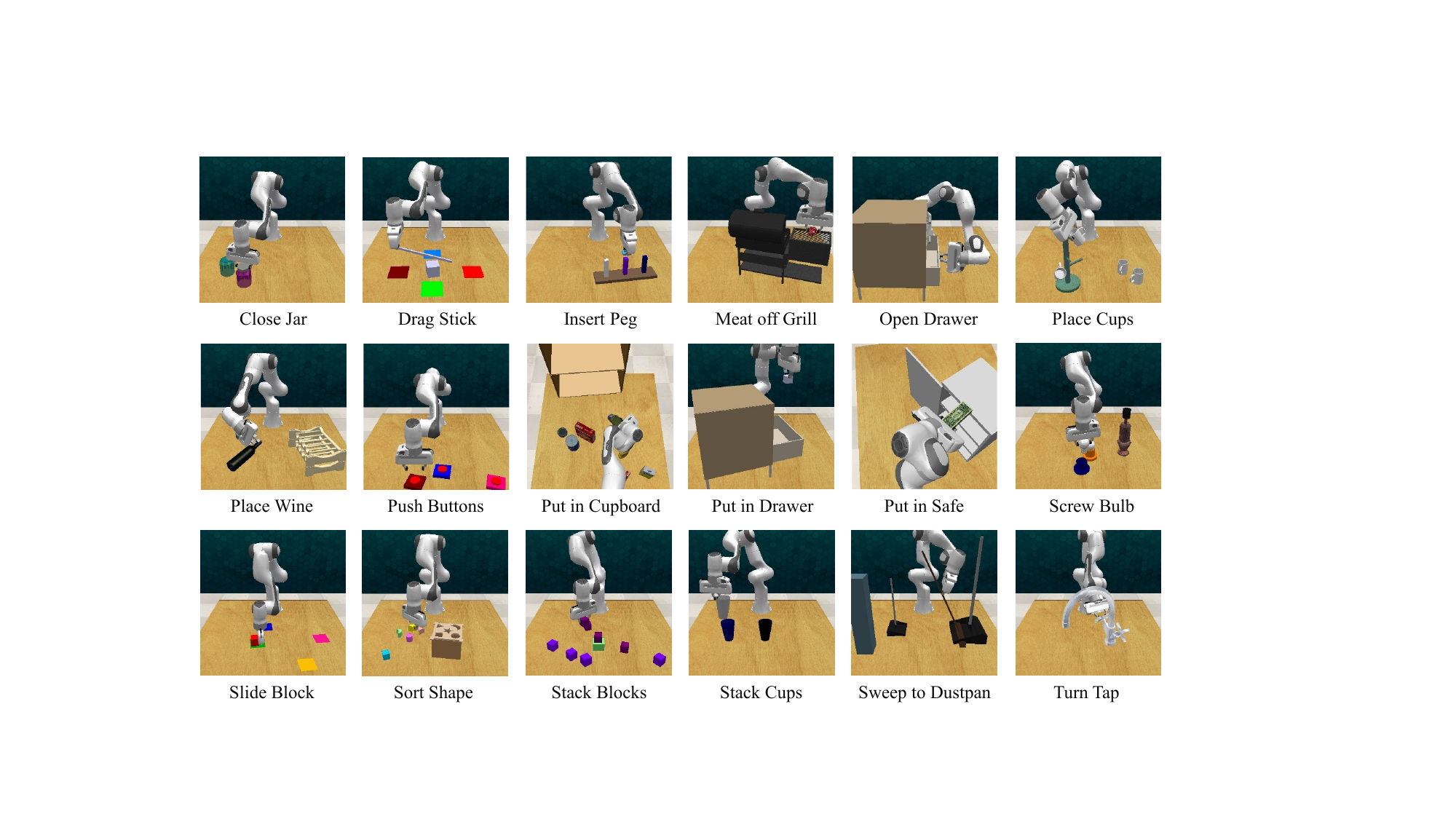}
    \caption{\textbf{RLBench Manipulation Tasks.} We evaluate~\method~on 18 simulated RLBench tasks, covering 249 variations of object poses, goal configurations, and scene appearances. During evaluation, the robot must complete each task within 25 execution steps under randomized colors, shapes, sizes, and semantic arrangements.}
    \label{fig:rlbench_tasks}
\end{figure*}

\begin{figure*}[t]
    \centering
    \includegraphics[width=1\linewidth]{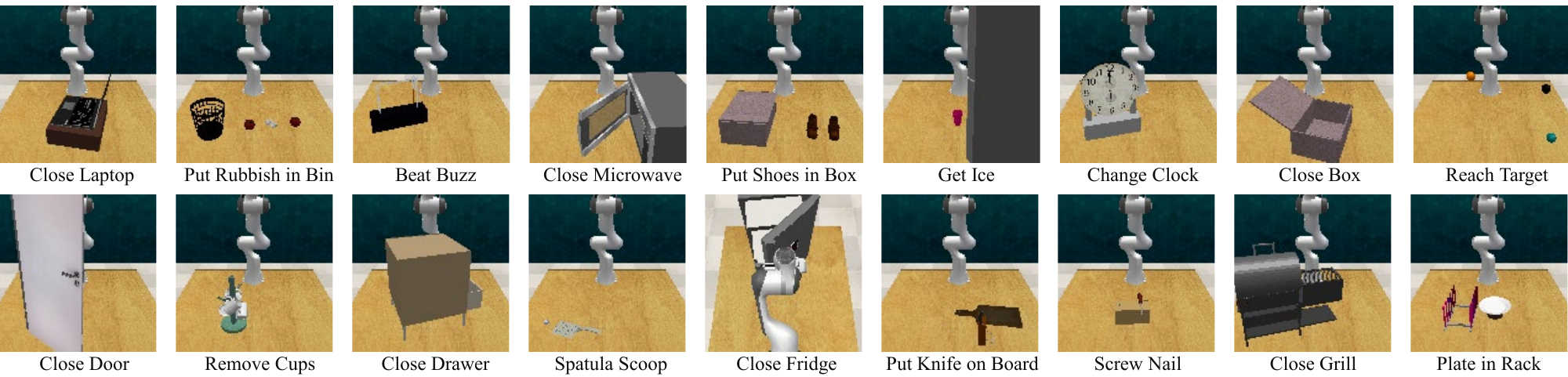}
    \caption{\textbf{Few-Shot tasks.} We adapt pre-trained model to 19 new tasks using only 10 demonstrations per task (1/10th of original data).}
    \label{fig:few_shot_tasks}
\end{figure*}

We follow the evaluation protocol of PerAct and benchmark~\method~on the same 18 RLBench tasks shown in Figure~\ref{fig:rlbench_tasks}. 
In total, these tasks cover 249 randomized scene configurations varying in object color, count, placement, shape, size, or category. Table~\ref{tab:rlbench_tasks} provides each task’s language instruction template, the number of variations, and the type of variation. 
Furthermore, Figure~\ref{fig:few_shot_tasks} illustrates the simulation tasks in a few-shot setting. 

\section{Real-world Tasks}

\begin{figure*}[h]
    \centering
    \includegraphics[width=0.7\linewidth]{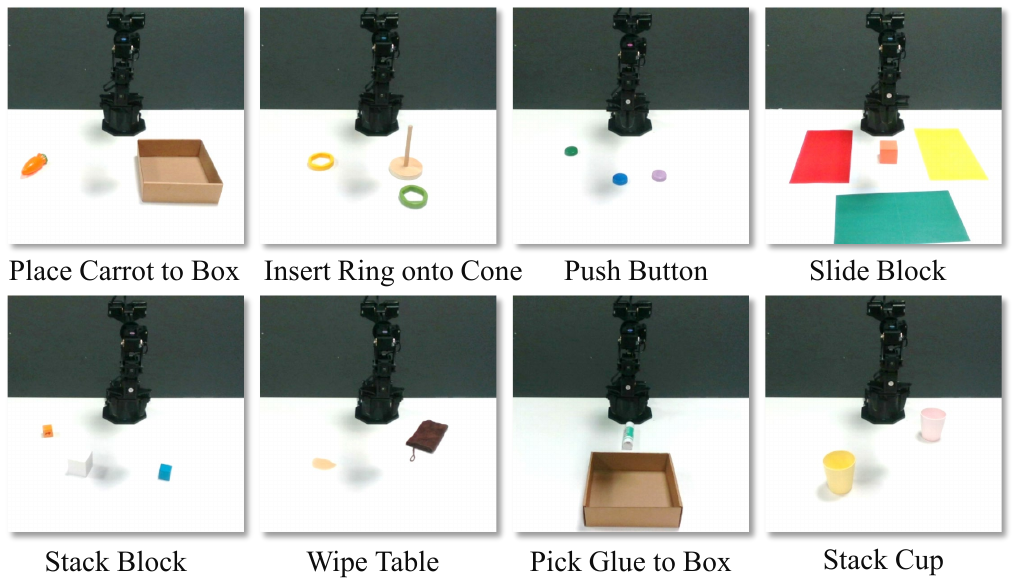}
    \caption{\textbf{Real-world tasks.} We employed 8 distinct tasks with a total of 15 variants in real-world experiments. }
    \label{fig:real_tasks_supp}
\end{figure*}

We evaluate~\method~on 8 real-world manipulation tasks illustrated in Figure \ref{fig:real_tasks_supp}. In total, these tasks comprise 15 randomized variants. Table \ref{tab:real_tasks} lists each task’s language instruction template and the corresponding number of variants. 

\begin{table}[t]
    \centering
    \begin{tabular}{l l}
        \toprule
        Hyperparameters            & Value\\
        \midrule
        batch size                 & $24 \times 8$\\
        learning rate              & 2.4e-3 \\
        optimizer                  & LAMB \\
        learning rate schedule     & cosine decay \\
        warmup steps               & 2000 \\
        training steps             & 40k\\
        training epochs            & 50\\
        \bottomrule
    \end{tabular}
    \caption{\textbf{Hyperparameter Settings for Training.}}
    \label{tab:hyperparameters}
\end{table}

\begin{table*}[h]
    \centering
    \resizebox{1.0\linewidth}{!}{%
    \begin{tabular}{c|ccccccccc}
    \toprule
       Samples  &  Avg. & Open Drawer & Put in Cupboard & Sort Shape & Stack Blocks & Place Cups & Screw Bulb & Stack Cups & Insert Peg
        \\
        \midrule
        25 &69.3 &74.0 &72.0 &62.0 &52.0 &52.0 &82.0 &74.0 &86.0 \\
        50 & 73.0 &81.3 &74.7 &62.7 &54.7 &52.0 &92.0 &77.3 &89.3 \\
        100 & 75.4 &82.0 &72.0 &73.3 &56.0 &56.7 &88.7 &81.3 &93.3 \\
        200 & 80.0 &82.7 &74.7 &72.0 &72.0 &64.0 &94.7 &80.0 &100.0 \\
        500 & 81.3 &86.7 &78.7 &73.3 &77.3 &78.7 &92.0 &64.0 &100.0 \\
        \bottomrule
    \end{tabular}}
    \caption{\textbf{Data Scaling Analysis. } 
Success rates (\%) of our model on eight challenging RLBench manipulation tasks as training sample size increases. The average success rate increases steadily from 69.3 \% at 25 samples to 73.0 \% at 50 samples, 75.4 \% at 100 samples and 80.0 \% at 200 samples, reaching 81.3 \% at 500 samples. These results demonstrate clear scaling behavior: adding more training data yields consistent gains in both overall and task-specific performance.}
    \label{tab:data_scaling}
\end{table*}

\begin{table*}[h!]
    \centering
    \resizebox{\textwidth}{!}{%
    \begin{tabular}{l*{10}{c}}
        \toprule
        Model 
        & \multicolumn{2}{c}{Avg.\ Success} 
        & Close Jar 
        & Drag Stick 
        & Insert Peg 
        & Meat off Grill 
        & Open Drawer 
        & Place Cups 
        & Place Wine 
        & Push Buttons \\
        \midrule
        Single-view 
            & \multicolumn{2}{c}{80} 
            & 92 
            & 100 
            & 64 
            & 100 
            & 76 
            & 32 
            & 100 
            & 100 \\
        Multi-view (ours)
            & \multicolumn{2}{c}{87.4} 
            & 94 
            & 100 
            & 93.3 
            & 98.7 
            & 82 
            & 56.7 
            & 94.7 
            & 100 \\
        \bottomrule
    \end{tabular}}
    \vspace{0.051em}
    \resizebox{\textwidth}{!}{%
    \begin{tabular}{l*{10}{c}}
        \toprule
        Model 
        & Put in Cupboard 
        & Put in Drawer 
        & Put in Safe 
        & Screw Bulb 
        & Slide Block 
        & Sort Shape 
        & Stack Blocks 
        & Stack Cups 
        & Sweep to Dustpan 
        & Turn Tap \\
        \midrule
        Single-view 
            & 48 
            & 84 
            & 96 
            & 84 
            & 80 
            & 64 
            & 68 
            & 56 
            & 100 
            & 96 \\
        Multi-view (ours)
            & 72 
            & 98.7 
            & 96.7 
            & 88.7 
            & 91.3 
            & 73.3 
            & 56 
            & 81.3 
            & 100 
            & 95.3 \\
        \bottomrule
    \end{tabular}}
    \caption{\textbf{Comparison of Single-view and Multi-view Setup.} Single-view uses front camera only, while multi-view combines front, left shoulder, and right shoulder views. The multi-view model raises average success from 80.0\% to 87.4\% on 18 RLBench manipulation tasks, demonstrating the benefits of multiple viewpoints. }
    \label{tab:view}
\end{table*}

\begin{table*}[h!]
    \centering
    \resizebox{\textwidth}{!}{%
    \begin{tabular}{l*{10}{c}}
        \toprule
        Model 
        & \multicolumn{2}{c}{Avg.\ Success} 
        & Close Jar 
        & Drag Stick 
        & Insert Peg 
        & Meat off Grill 
        & Open Drawer 
        & Place Cups 
        & Place Wine 
        & Push Buttons \\
        \midrule
        DINO 
            & \multicolumn{2}{c}{86.5}
            & 94.7
            & 96
            & 96
            & 97.3
            & 80
            & 52
            & 93.3
            & 100 \\
        CLIP (ours)
            & \multicolumn{2}{c}{87.4} 
            & 94 
            & 100 
            & 93.3 
            & 98.7 
            & 82 
            & 56.7 
            & 94.7 
            & 100 \\
        \bottomrule
    \end{tabular}}
    \vspace{0.051em}
    \resizebox{\textwidth}{!}{%
    \begin{tabular}{l*{10}{c}}
        \toprule
        Model 
        & Put in Cupboard 
        & Put in Drawer 
        & Put in Safe 
        & Screw Bulb 
        & Slide Block 
        & Sort Shape 
        & Stack Blocks 
        & Stack Cups 
        & Sweep to Dustpan 
        & Turn Tap \\
        \midrule
        DINO 
            & 76
            & 98.7
            & 100
            & 88
            & 80
            & 69.3
            & 61.3
            & 84
            & 98.7
            & 92 \\
        CLIP (ours)
            & 72 
            & 98.7 
            & 96.7 
            & 88.7 
            & 91.3 
            & 73.3 
            & 56 
            & 81.3 
            & 100 
            & 95.3 \\
        \bottomrule
    \end{tabular}}
    \caption{\textbf{Ablation of Visual Backbone.} We compare DINO and CLIP as visual backbones on 18 RLBench manipulation tasks. The CLIP backbone raises average success from 86.5\% to 87.4\%, demonstrating the benefit of language-aligned visual features.}
    \label{tab:backbone}
\end{table*}

\begin{table*}[h!]
    \centering
    \resizebox{\textwidth}{!}{%
    \begin{tabular}{l p{7cm} c l}
        \toprule
        Task Name       & Language Instruction Template                 & \# Variations & Variation Type   \\
        \midrule
        Close Jar       & “close the [ ] jar”                          & 20            & color            \\
        Drag Stick      & “use the stick to drag the cube onto the [ ] target” & 20    & color            \\
        Insert Peg      & “put the ring on the [ ] spoke”               & 20            & color            \\
        Meat off Grill  & “take the [ ] off the grill”                  & 2             & category         \\
        Open Drawer     & “open the [ ] drawer”                         & 3             & placement        \\
        Place Cups      & “place [ ] cups on the cup holder”            & 3             & count            \\
        Place Wine      & “stack the wine bottle to the [ ] of the rack” & 3            & placement        \\
        Push Buttons    & “push the [ ] button, then the [ ] button”    & 50            & color            \\
        Put in Cupboard & “put the [ ] in the cupboard”                 & 9             & category         \\
        Put in Drawer   & “put the item in the [ ] drawer”              & 3             & placement        \\
        Put in Safe     & “put the money away in the safe on the [ ] shelf” & 3         & placement        \\
        Screw Bulb      & “screw in the [ ] light bulb”                 & 20            & color            \\
        Slide Block     & “slide the block to the [ ] target”           & 4             & color            \\
        Sort Shape      & “put the [ ] in the shape sorter”             & 5             & shape            \\
        Stack Blocks    & “stack [ ] blocks”                            & 60            & color, count     \\
        Stack Cups      & “stack the other cups on top of the [ ] cup”  & 20            & color            \\
        Sweep to Dustpan& “sweep dirt to the [ ] dustpan”               & 2             & size             \\
        Turn Tap        & “turn [ ] tap”                                & 2             & placement        \\
        \bottomrule
    \end{tabular}}
    \caption{\textbf{RLBench Task Set.} 18 manipulation tasks with corresponding language instruction templates, number of randomized variations, and variation types (color, count, placement, shape, size, category), totaling 249 distinct scene configurations.}
    \label{tab:rlbench_tasks}
\end{table*}

\begin{table*}[h]
    \centering
    \resizebox{0.85\textwidth}{!}{%
    \begin{tabular}{l p{7cm} c }
        \toprule
        Task Name               & Language Instruction Template                    & \# Variations \\
        \midrule
        Place Carrot To Box     & “place the carrot into the box”                  & 1             \\
        Insert Ring Onto Cone   & “insert the [] ring onto the cone”& 2             \\
        Push Button             & “push the [ ] button”                            & 3             \\
        Slide Block             & “slide the block to the [ ] target”              & 3             \\
        Stack Block             & “stack the [ ] block on the other block”& 2             \\
        Wipe Table              & “wipe the table”                                 & 1             \\
        Pick Glue To Box        & “pick up the glue stick and place it in the box” & 1             \\
        Stack Cup               & “stack the [ ] cup on the other cup”& 2             \\
        \bottomrule
    \end{tabular}}
    \caption{\textbf{Real-World Task Set.} 8 real-world tasks with their language instruction templates and number of variations.}
    \label{tab:real_tasks}
\end{table*}

\end{document}